\documentclass[conference]{IEEEtran}
\IEEEoverridecommandlockouts
\usepackage{cite}
\usepackage{amsmath,amssymb,amsfonts}
\usepackage{algorithmic}
\usepackage{graphicx}
\usepackage{textcomp}
\usepackage{xcolor}
\usepackage{marvosym} 
\usepackage{booktabs} 
\usepackage{tabularx}
\usepackage{url}
\def\BibTeX{{\rm B\kern-.05em{\sc i\kern-.025em b}\kern-.08em
    T\kern-.1667em\lower.7ex\hbox{E}\kern-.125emX}}
\begin{document}

\title{Segment-Anything Models Achieve Zero-shot Robustness in Autonomous Driving}

\author{Jun Yan$^{1}$, Pengyu Wang$^{1}$, Danni Wang$^{1}$, Weiquan Huang$^{1}$, Daniel Watzenig$^{2}$, and Huilin Yin$^{1}$\thanks{$^{1}$Jun Yan, Pengyu Wang, Danni Wang, Weiquan Huang, and Huilin Yin (\texttt{yanjun@tongji.edu.cn, perry0817@tongji.edu.cn, danni\_Wang0328@163.com, weiquanh@tongji.edu.cn, yinhuilin@tongji.edu.cn}) are with the College of Electronics and Information Engineering, Tongji University, Shanghai 201804, China.}\thanks{$^{2}$Daniel Watzenig (\texttt{daniel.watzenig@tugraz.at}) is with the Institute of Electrical Measurement and Sensor Systems, Graz University of Technology and with the Virtual Vehicle Research, Graz 8010, Austria.}%
\thanks{Huilin Yin is the corresponding author \textsuperscript{\Letter}.}
}

\maketitle

\begin{abstract}
Semantic segmentation is a significant perception task in autonomous driving. It suffers from the risks of adversarial examples. In the past few years, deep learning has gradually transitioned from convolutional neural network (CNN) models with a relatively small number of parameters to foundation models with a huge number of parameters. The segment-anything model (SAM) is a generalized image segmentation framework that is capable of handling various types of images and is able to recognize and segment arbitrary objects in an image without the need to train on a specific object. It is a unified model that can handle diverse downstream tasks, including semantic segmentation, object detection, and tracking. In the task of semantic segmentation for autonomous driving, it is significant to study the zero-shot adversarial robustness of SAM. Therefore, we deliver a systematic empirical study on the robustness of SAM without additional training. Based on the experimental results, the zero-shot adversarial robustness of the SAM under the black-box corruptions and white-box adversarial attacks is acceptable, even without the need for additional training. The finding of this study is insightful in that the gigantic model parameters and huge amounts of training data lead to the phenomenon of emergence, which builds a guarantee of adversarial robustness. SAM is a vision foundation model that can be regarded as an early prototype of an artificial general intelligence (AGI) pipeline. In such a pipeline, a unified model can handle diverse tasks. Therefore, this research not only inspects the impact of vision foundation models on safe autonomous driving but also provides a perspective on developing trustworthy AGI. The code is available at: \url{https://github.com/momo1986/robust_sam_iv}.
\end{abstract}


\section{Introduction}
Semantic segmentation is a significant task in autonomous driving. It helps realize the environmental perception, path planning and decision, transportation scenario analysis, barrier avoidance and collision prevention, precise localization, and human-computer interaction based on the visualization interface. The performance of semantic segmentation guarantees the Safety of the Intended Functionality (SOTIF) in autonomous driving. Over the past decade, with the successful application of deep learning, many semantic segmentation models~\cite{FCN,Segnet,Pspnet,Deeplabv3+,SegFormer,ISANet,OCRNet,STDC} have become well known.
 \begin{figure*}[!t]
        \centering
\includegraphics[width=0.85\hsize]{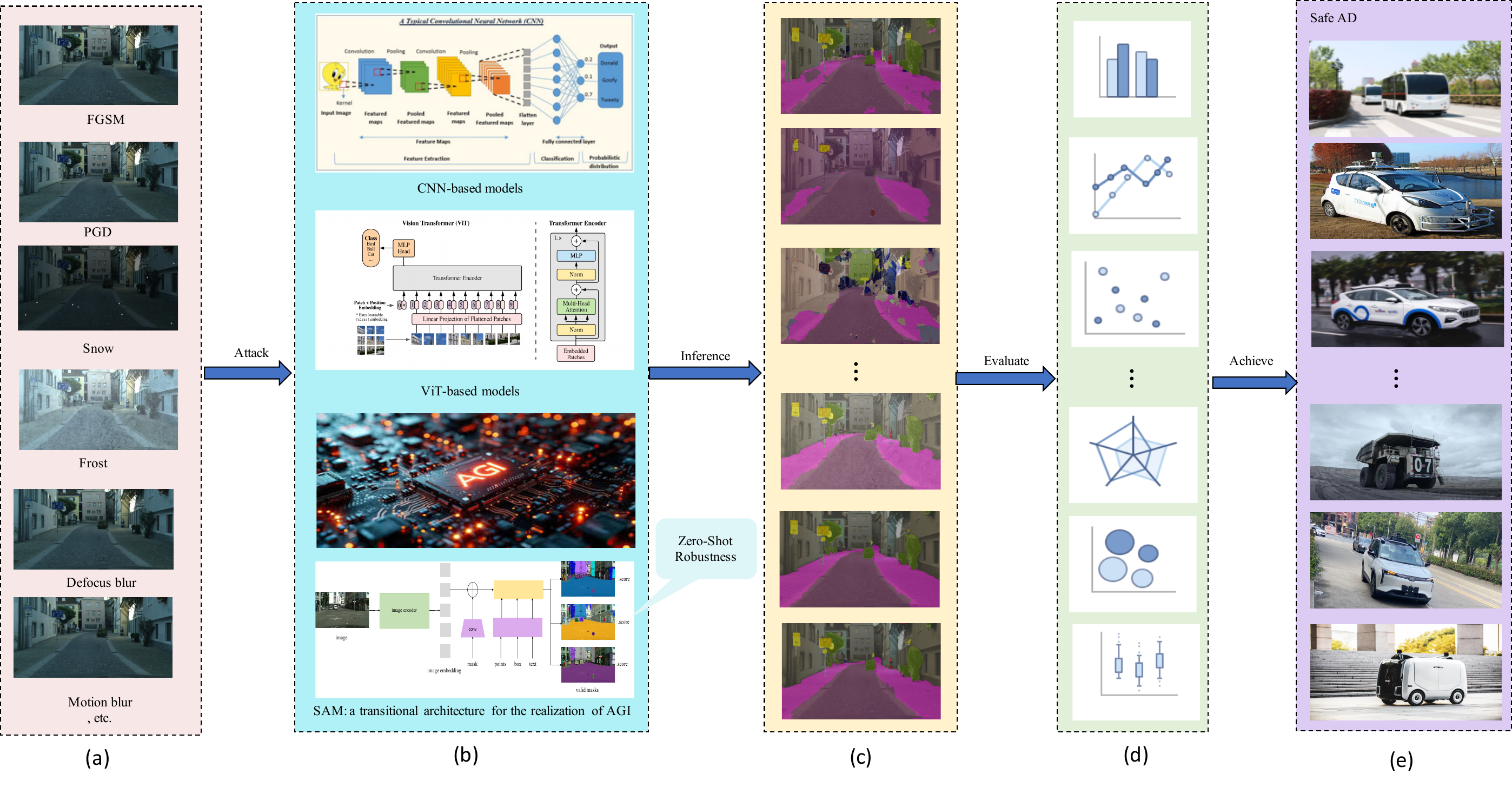}
\caption{Framework of robustness study on SAM. The drawing material of CNN is credit to the webpage~\cite{cnn_fig}, and the drwaing material of ViT is credit to the paper~\cite{ViT}.}
\label{fig:advRobustSemFramework}
\end{figure*}
\par The existence of adversarial examples~\cite{FGSM,PGD,GDUAP} is a huge challenge towards the trustworthy deep learning. The tiny perturbations would not change the semantic information of an image. These adversarial examples are imperceptible to the human eyes. However, they can deceive the neural networks to make wrong predictions. The application of adversarial samples in semantic segmentation tasks can cause confusion in the classification of different pixels. This fault cannot be tolerated in autonomous driving which would cause a potential security risk.
\par It raises an insightful scientific problem. Which adversarial sample has the highest security risk against semantic segmentation models? What kind of models are most robust under adversarial attacks? Is it possible to utilize the adversarial examples as responsible tools to assist in the safety testing of autonomous driving?
\par Previously, some research has covered the discussions on adversarial robustness of semantic segmentation models based on convolutional neural networks (CNNs)~\cite{SegmentationRobustnessICLR17,SegmentationRobustnessOxford,SegmentationRobustnessHeidelberg}. After the rise of vision transformers (ViTs) based on the aggregated tokens in the tasks of visual recognition, the research on the reliability of ViT-based semantic segmentation models including their adversarial robustness and performance under natural shifts~\cite{VITRobustness} shows its significance. Recently, it has been witnessed that the ViT model as a foundation model can perform various downstream tasks~\cite{SAM}. Concerns about their robustness make sense~\cite{SAMRobustness,SAMRobustness2}. Some empirical studies have given the specific attention to adversarial robustness of the specific deep-learning-based semantic segmentation models, either on CNN-based models\cite{SegmentationRobustnessICLR17,SegmentationRobustnessOxford,SegmentationRobustnessHeidelberg} or segment-anything model (SAM)~\cite{SAMRobustness2}. Nevertheless, the change in the model design paradigm brings about a change in the robustness study paradigm. The CNN-based models are sensitive to gradient-based attacks or corruption. Such security threats may no longer be so fatal to the SAM models.
\par The research assumption indicates the necessity of conducting a systematic study on the adversarial robustness of diverse model structures in semantic segmentation for autonomous driving. There is a wide variety of adversarial examples that may attack autonomous vehicles with the deployment of deep-learning-based semantic segmentation models. The white-box adversarial examples are usually related to cybersecurity risks in the Internet of Vehicles, and the black-box adversarial examples based on corruption can be utilized as tools for safety-bound testing in autonomous driving, which enables security risks to be transformed into responsible applications. The SOTIF~\cite{sotif} deals with the risks caused by the limitations of AI models. The black-box corruption can help generate different test cases in the SOTIF evaluation. Therefore, there is a lot of room for exploration in the robustness study for autonomous driving that has not been looked at in previous studies.
\par Moreover, the appearance of GPT-4~\cite{GPT-4} and Segment-Anything model (SAM)~\cite{SAM} demonstrates that the foundation models can solve complex problems and achieve the human-level performance with the unification of the language signals and visual signals. These early but imperfect sparks of artificial general intelligence (AGI) connect to the phenomenon of emergence~\cite{MoreIsDifferent,EmergenceNaturePhysics} that huge model parameters with huge amounts of training data can cause the phase transition of performance of the AI agents. The unification of visual signals and language signals leads to several trends: the realization of open-world visual recognition, clustering the raw image pixels with the prompts based on a generalized foundation model, and generalized visual encoding. It is worth studying the relationship of such a trend with adversarial robustness.
\par Based on the research motivation, we explore the zero-shot adversarial robustness under the white-box attacks and black-box attacks with a comprehensive empirical study. Fig. \ref{fig:advRobustSemFramework} illustrates the pipeline. We implement the robustness evaluation at the data level for autonomous driving on the Cityscapes dataset~\cite{cityscapes} with a quantitative and qualitative analysis of our experiment result. At the model level, the evaluated models include the typical CNN and ViT models and up-to-date SAM models. We are particularly interested in the zero-shot adversarial robustness performance of SAM models with the constraint of the language encoders, e.g., Contrastive Language-Image Pre-Training (CLIP). The evaluation of the zero-shot adversarial robustness helps design trustworthy models in the generative AI era. 
\par This paper has two contributions:
\begin{itemize}
\item (Methodology-wise) In the semantic segmentation task, this study shows a SAM pipeline with the assistance of text encoder achieves a robust in-context learning ability under the adversarial attacks. 
\item (Empirical-study-wise) We evaluate the robustness of CNN models, ViT models, and SAM models under the white-box attacks and black-box attacks on the dataset of Cityscapes~\cite{cityscapes}.
\end{itemize}
\section{Related Works}
This section gives a literature review on the semantic segmentation models based on deep learning, adversarial examples in the vision tasks.
\subsection{Deep-learning-based Semantic Segmentation Models}
The renaissance of deep learning in the vision task~\cite{VGG,Resnet,MobileNetV2,Inception} boosts the development of deep-learning-based semantic segmentation models. The first models to appear are based on CNNs, including Fully Convolutional Networks (FCN)~\cite{FCN}, SegNet~\cite{Segnet}, Pyramid Scene Parsing Network (PSPNet)~\cite{Pspnet}, DeepLabV3+ model~\cite{Deeplabv3+}, and so on. The recent study proposes the real-time Short-Term Dense Concatenate module (STDC module) to obtain the scalable receptive fields by progressively decreasing the dimension of the feature maps and extract the multi-scale information by connecting the response maps of multiple successive layers~\cite{STDC}.
\par ViT~\cite{ViT} uses a self-attention mechanism to capture global relational and contextual information in the input image rather than relying on the operation of local receptive fields of CNNs. By dividing the input image into a set of fixed-size blocks, the long-range image contextual information can be captured for feature extraction. Based on such an advantage, ViTs have become main-stream backbones in the task of semantic segmentation. SegFormer~\cite{SegFormer} incorporates a hierarchically structured Transformer encoder that produces multi-scale features, eliminating the need for positional encoding, and it depends on the multi-layer perception (MLP) decoder to aggregate the information from various layers. Object-contextual representations networks for semantic segmentation (OCRNet)~\cite{OCRNet} addresses the context aggregation problem in
semantic segmentation by exploring the potential to explicitly transform a pixel classification problem into an object region classification problem. Improved spatial attention network (ISANet)~\cite{ISANet} decomposes the original dense affinity matrix into two sparse affinity matrices to improve the efficiency of semantic segmentation model based on ViTs, where one sparse affinity matrix is used for long-distance transmission and the other sparse affinity matrix is used for short-distance transmission. One meaningful attempt is to unify different tasks in a single unique model, e.g., panoptic segmentation, instance segmentation, semantic segmentation~\cite{OneFormer}.
\par ViTs usually require more parameters. Therefore, such models are more suitable for large-scale datasets and more adequate computational resources compared with CNNs. It brings a new paradigm and trend in computer vision: pre-training on large-scale datasets to model the distributions of visual objects all over the world, unification of different granularities and tasks, and incorporation of other human knowledge such as languages~\cite{CLIP}. The model trained on the general dataset at scale and adapted to diverse downstream visual tasks is a foundation model~\cite{ReviewFoundationModels}. Recent milestone research has built the largest segmentation dataset to date, with over 1 billion masks on 11 million licensed and privacy-respecting images. The SAM model~\cite{SAM} and its lightweight variants~\cite{FastSAM,MobileSAM} are designed and trained to be promptable so that their zero-shot recognition performance is considerable.
\subsection{Adversarial Examples and Its Applications on Semantic Segmentation Models}
\par Deep learning models are vulnerable to adversarial examples that are imperceptible to human eyes. The carefully crafted adversarial perturbations to the input image can cause a neural network model to make incorrect predictions. Adversaries can obtain the gradient information via the model leakage, including fast gradient sign method (FGSM)~\cite{FGSM},
project gradient descent (PGD) attacks~\cite{PGD}, Carlini and Wagner attacks
(C\&W)~\cite{C&W}, and so on. Besides the model-specific and white-box attacks, there exist universal adversarial perturbations (UAPs) that the small, subtle perturbations can be applied to a wide range of input data and deceive diverse neural networks with a good transferability~\cite{UAP,GDUAP}.
\par Sometimes, the attack scenarios are model-agnostic that the attackers without the model knowledge try to generate the adversarial examples via the iteration search~\cite{NES,GeoDA}. The dense adversary generation (DAG) attack method~\cite{DAG} is a typical black-box method to disturb the semantic segmentation models. The black-box adversarial examples can be both model-agnostic and data-agnostic to fool the neural networks via the procedural noise functions~\cite{image_corruptions_origin}. When these corruptions are applied to autonomous driving, it transfers the security issues to the safety testing scenarios~\cite{image_corruptions}.
\par There are several empirical studies on the adversarial attacks against the semantic segmentation models in autonomous driving. Some early work focuses on the specific CNN models~\cite{SegmentationRobustnessICLR17,SegmentationRobustnessOxford,SegmentationRobustnessHeidelberg}. Although the study to cover more attack scenarios and model structures cannot be neglected~\cite{OurIVPaper}, it is still insufficient for current research topics on foundation models. Moreover, some research on adversarial the robustness of SAM does not focus on autonomous driving~\cite{SAMRobustness,SAMRobustness2}, which reflects the necessity for this study. Shan et al.~\cite{shan2023robustness} provides a preliminary evaluation of the robustness of SAM in adverse weather. However, a systematic assessment including both white-box and black-box attacks is still needed to connect the empirical study to SOTIF and cybersecurity in autonomous driving. 

\section{SAM Based on the Open-set Category Encoder}
 \begin{figure}[!t]
        \centering
\includegraphics[width=0.95\hsize]{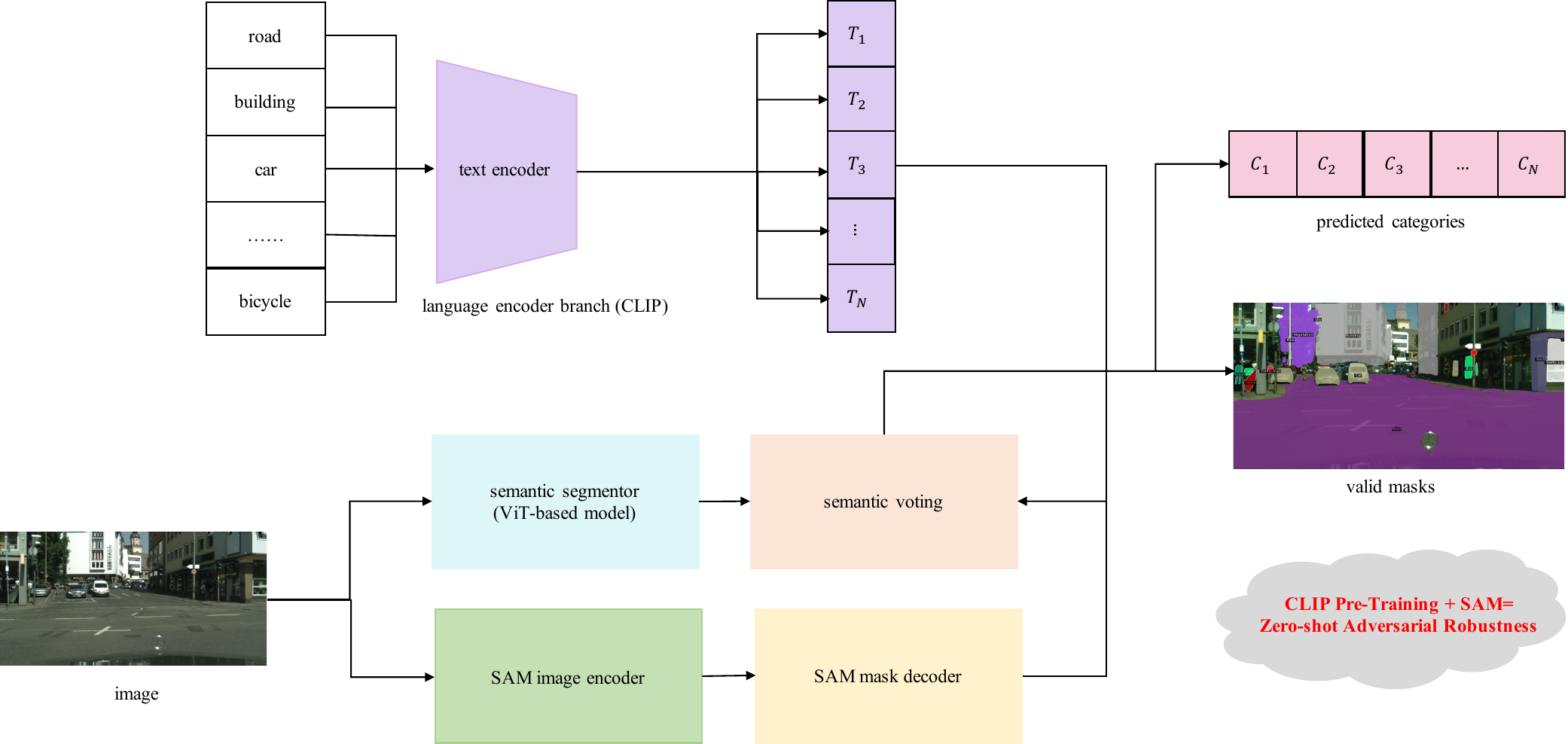}
\caption{The framework of the Semantic-Segment-Anything model (SSAM) consists of SAM and open-set category encoder }
\label{fig:sam_clip}
\end{figure}
\par SAM~\cite{SAM} is a powerful foundation model that can segment arbitrary objects, and SA-1B is the largest segmentation dataset to approximate the world distribution with 11M images. SAM is a generalizable object segmentation method that delivers precise contouring through its masks. SA-1B emerges as the large-scale generalized segmentation dataset. The closed-set  semantic segmentation pipeline based on deep learning can provide rich semantic annotations, while SAM can generate precise masks. Moreover, the CLIP model~\cite{CLIP} is a large multimodal pretraining model. CLIP can map images and text into a common feature space to form an image-text pair. CLIP contains two main components: an image encoder and a text encoder. The two encoders are trained in parallel to achieve cross-modal feature representation and alignment through contrast learning.
\par This study raises the question of whether the zero-shot adversarial robustness when SAM meets the open-set category open-set category encoder based on the CLIP method. Fig. \ref{fig:sam_clip} describes such a framework. The SAM image encoder (green part) converts raw image data into low-dimensional feature vectors for subsequent computation and analysis. The role of the mask decoder is to efficiently map image embeddings, prompt embeddings, and output markers to masks. It uses a modified Transformer decoder block followed by a dynamic mask prediction header. The mask decoder (yellow part) upsamples the image embeddings and uses MLP to map the output markers to a dynamic linear classifier that will compute the mask foreground probability for each image location. The method of CLIP~\cite{CLIP} helps categorize different classes in different application scenarios (e.g., 19 categories on the Cityscapes dataset~\cite{cityscapes}). In Fig. \ref{fig:sam_clip}, the purple part represents the text encoder based on the CLIP approach. CLIP has rich world knowledge, reasoning ability to pair with images, and can assist in the segmentation branch. The semantic branch (blue part) provides per-pixel categories, implemented by a semantic segmentation pipeline, which can be customized by the user with the ViT-based model (e.g., SegFormer~\cite{SegFormer} and OneFormer~\cite{OneFormer}) according to the architecture of the segmentation model and the categories of interest. \textit{It deserves to be emphasized that the Semantic-Segment-Anything models do not need supervised learning of SegFormer and OneFormer on the specific dataset. Since SAM is a visual foundation model, it can transfer to the downstream task without additional training.} The semantic voting module (orange part) crops the corresponding pixel categories based on the position of the mask. The framework selects the top-1 predicted category of these pixels as the categorization result of this mask.
\par SAM is based on the ViT model, which can express image information exclusively as a token. CLIP generates linguistic commands that can help to extract visual information. Based on the combination of image and language, this paradigm can deal with complex scenes and realize the zero-shot adversarial robustness of the visual perception system in the downstream tasks.
\par It leads to the central theme of this study: what kind of model architecture can guarantee the adversarial robustness of semantic segmentation models based on neural networks? CNN-based segmentation models have better generalization, but their adversarial robustness needs to be improved. ViT model extracts better global information about image objects, eliminating the inductive bias present in CNN models and their sensitivities to textures~\cite{CNNTextureBias}. At the same time, the Transformer architecture creates a potential to unify the image and text information. It raises the question of whether we can achieve better adversarial robustness with the utilization of a ViT-based foundation model combined with CLIP assistance for the task of semantic segmentation in autonomous driving. Therefore, a comprehensive and systematic empirical study is necessary.
\section{Experiment}
\subsection{Experiment Settings}
\subsubsection{Dataset}
The Cityscapes dataset~\cite{cityscapes} is a publicly available dataset used for studying semantic segmentation and scene understanding vision tasks. This dataset contains high-resolution ($1024 \times 2048$) images from different cities in Germany and Switzerland that capture streets, buildings, vehicles, pedestrians, and other elements of the urban environment. It contains the categories of road, sidewalk, building, wall, fence, pole, traffic light, traffic sign, vegetation, terrain, sky, person, rider, car, truck, bus, train, motorcycle, and bicycle. This study mainly focuses on the scientific problem in the inference time. Thus, the experiments are implemented on the validation dataset with $500$ images.
\subsubsection{Evaluation Metrics}
The semantic segmentation task mainly uses recall, precision, F1-score, and intersection-overunion (IoU) metrics to evaluate the performance of models. The metric of mIoU denotes the IoU scores from $N$ pairs of data samples:
\begin{equation}\label{eq:miou}
m I o U=\frac{1}{N} \sum_{i=1}^{N} \frac{\bigcap\left(\boldsymbol{m a s k}_{\text {groundtruth }}, \boldsymbol{m a s k}_{\text {predict }}\right)}{\left.\bigcup \boldsymbol { (mask }_{\text {groundtruth }}, \boldsymbol{m a s k}_{\text {predict }}\right)}
\end{equation}
In Eq.\ref{eq:miou}, $\operatorname{mask}_{(\cdot)}$ denotes a binary matrix of categorized pixels, $\boldsymbol{m a s k}_{\text {groundtruth }}$ and $\boldsymbol{m a s k}_{\text {predict}}$ are the masked region of content images and the predicted masked regions, respectively. $\bigcap$ and $\bigcup$ denote the intersection and union section, respectively. The mIoU score ranges from $[0.0,\ 1.0]$. The higher mIoU score represents considerable robustness performance.
\subsubsection{Models}
The CNN-based segmentation models include FCNs~\cite{FCN}, DeepLabV3+ (Abbreviation ``DP") models~\cite{Deeplabv3+}, SegNet models~\cite{Segnet}, and PSP-Net models~\cite{Pspnet}. The symbols of FCN32s, FCN16s, and FCN8s represent the FCN models with the strategies of 32 times upsampling, 16 times upsampling, and 8 times upsampling. The backbones of semantic segmentation models are pre-trained on the ImageNet dataset~\cite{ImageNet}, including ResNet~\cite{Resnet}, VGG16~\cite{VGG}, MobileNetV2~\cite{MobileNetV2}, Xception~\cite{xception}, and High-Resolution Network (HRNet)~\cite{HRNet}. The ViT-based segmentation models include Segformer~\cite{SegFormer}, OCRNet~\cite{OCRNet}, STDC~\cite{STDC}, ISANet~\cite{ISANet}, and OneFormer~\cite{OneFormer}. All of the models are implemented in the PyTorch framework. This study selects SAM as a foundation model. The pipeline combined with language encoder and SAM is refined based on an open-source repository~\cite{Semantic-Segment-Anything}. The study covers two types of SAM architectures: vanilla SAM~\cite{SAM} and Mobile-SAM~\cite{MobileSAM}. There are two different backbones: SegFormer~\cite{SegFormer} and OneFormer~\cite{OneFormer}. The exploration of the difference between the Transformer model with a single segmentation task and the Transformer model with the unification of multiple tasks is insightful.
\subsubsection{Adversarial Attack Methods}
The white-box adversarial attacks mean that the attackers fully understand the internal structures and parameters of the machine learning models, while black-box attacks mean that the attackers only know the inputs and outputs of the models and need to infer the internal structures and parameters of the models by observing the responses of the models. The white-box attack methods include FGSM~\cite{FGSM}, PGD~\cite{PGD}. The black-box image corruptions~\cite{image_corruptions_origin} are evaluated in this study. The black-box attacks can simulate the abnormal weather and camera distortions. The image-corruption perturbations can be generated with 19 different procedural adversarial noises, including Gaussian noise, shot noise, impulse noise, defocus blur, glass blur, motion blur, zoom blur, snow, frost, fog, brightness, contrast, elastic transform, pixelate, JPEG compression, speckle noise, Gaussian blur, spatter, saturate. Each kind of adversarial noise has 5 severity levels. The vulnerabilities of semantic segmentation models under such black-box corruptions can be attributed to the risk issues of SOTIF.
\subsection{Main Experimental Results}
\subsubsection{Robustness Study of Black-box Corruptions}
\begin{table*}[t]\scriptsize
\centering
\setlength\tabcolsep{2pt}
\caption{Robustness of semantic segmentation models under the black-box corruptions (Severity=1-3)}
\label{table:black_box_corruption}
\resizebox{\textwidth}{!}{

\begin{tabular}{@{}l|c|ccccc|cccc|cccc|cccccc@{}}
\toprule
                 &      & \multicolumn{5}{c|}{\textbf{Blur}}         & \multicolumn{4}{c|}{\textbf{Weather}} & \multicolumn{4}{c|}{\textbf{Noise}} & \multicolumn{6}{c}{\textbf{Digital}}                        \\ \midrule
                 Architecture & \textbf{clean} & Gaussian & \textbf{defocus} & \textbf{motion} & glass & zoom & \textbf{snow}   & \textbf{frost}   & \textbf{fog}    & \textbf{spatter}   & speckle & Gaussian & shot & impulse & brightness & contrast & JPEG & saturate & pixelate & elastic \\ \midrule
DP-ResNet50~\cite{Deeplabv3+}      & 76.6      & 66.0     & 62.2    & 65.3   & 52.9  & 24.3 & 19.6   & 33.7    & 72.5   & 45.3      & 32.8    & 10.7     & 14.0 & 9.5     & 74.4       & 70.6     & 34.0 & 74.6     & 63.3     & 74.3    \\
DP-ResNet101~\cite{Deeplabv3+}      & 77.3      & 67.6     & 64.9    & 66.4   & 54.2  & 26.8 & 24.9   & 37.9    & 74.0   & 52.3      & 41.2    & 15.6     & 20.2 & 14.3    & 75.1       & 73.3     & 40.4 & 75.5     & 66.6     & 74.6    \\
DP-MobileNetV2~\cite{Deeplabv3+}    & 72.9      & 59.6     & 55.8    & 60.0   & 44.6  & 21.8 & 19.6   & 28.1    & 66.9   & 47.8      & 26.0    & 9.6      & 10.6 & 12.1    & 69.4       & 64.3     & 22.1 & 69.8     & 54.7     & 71.1    \\
DP-Xception65~\cite{Deeplabv3+}    & 78.4      & 71.6     & 68.6    & 69.6   & 62.5  & 25.0 & 19.9   & 26.5    & 65.8   & 60.8      & 28.4    & 7.0      & 8.0  & 5.1     & 66.6       & 71.9     & 34.1 & 67.1     & 71.1     & 76.2    \\ \midrule
FCN-ResNet50~\cite{FCN}     & 67.9      & 55.9     & 52.5    & 56.5   & 43.4  & 22.6 & 17.3   & 31.1   & 48.6   & 45.2      & 10.1    & 1.8      & 2.2  & 1.5     & 54.6       & 60.4     & 15.4 & 52.0     & 33.6     & 64.8    \\
FCN-ResNet101~\cite{FCN}     & 69.0      & 57.3     & 54.1   & 56.6   & 43.5  & 21.9 & 17.3   & 27.9    & 53.4   & 47.1      & 17.0    & 4.8      & 5.1  & 6.0     & 58.0       & 59.6     & 22.9 & 52.8     & 39.8     & 65.9    \\
FCN32s-VGG16~\cite{FCN}      & 55.1      & 42.5     & 38.9    & 41.6   & 28.7  & 17.9 & 12.9   & 18.8    & 34.8   & 36.8      & 14.0    & 7.7      & 8.8  & 2.7     & 40.0       & 41.7     & 15.5 & 38.2     & 22.1     & 52.7    \\
FCN16s-VGG16~\cite{FCN}      & 58.4      & 42.9     & 38.4    & 42.9   & 29.8  & 18.0 & 12.6   & 17.9    & 33.5   & 37.7      & 8.6     & 2.4      & 3.3  & 2.3     & 40.9       & 44.0     & 14.3 & 39.3     & 21.9     & 56.1    \\
FCN8s-VGG16~\cite{FCN}       & 60.3      & 44.3     & 40.1    & 43.4   & 32.2  & 18.0 & 12.2   & 16.9    & 34.0   & 39.4      & 8.6     & 3.2      & 3.8  & 2.3     & 42.8       & 43.6     & 15.4 & 41.9     & 22.4     & 57.5    \\ \midrule
PSPNet-ResNet50~\cite{Pspnet}  & 69.3      & 57.1     & 54.1    & 57.4   & 35.4  & 22.8 & 10.3   & 17.8    & 45.0   & 42.9      & 9.5     & 2.8      & 3.1  & 4.6     & 56.8       & 56.1     & 13.9 & 49.0     & 25.9     & 65.6    \\
PSPNet-ResNet101~\cite{Pspnet} & 70.7      & 58.8     & 55.4    & 59.0   & 43.8  & 24.6 & 15.4   & 32.7    & 54.9   & 44.1      & 12.9    & 3.4      & 4.5  & 3.5     & 60.3       & 60.9     & 27.1 & 53.6     & 43.6     & 66.9    \\
SegNet-VGG16~\cite{Segnet}     & 62.7      & 52.4     & 49.0    & 54.8   & 51.8  & 23.4 & 18.5   & 24.9    & 42.4   & 49.5      & 40.2    & 18.4     & 22.8 & 13.0    & 56.2       & 52.9     & 37.7 & 52.9     & 62.0     & 61.5    \\ \midrule
OCRNet-ResNet100~\cite{OCRNet} & 80.2 & 68.9 & 68.9 & 69.0 & 56.3 & 24.1 & 21.4 & 43.4 & 76.3 & 56.6 & 31.7 & 6.4 & 9.6& 12.9 & 79.1 & 73.4 & 28.5 & 78.6 & 69.3 & 78.4 \\
OCRNet-HRNet-W48~\cite{OCRNet} & 80.5 & 70.7 & 70.3 & 71.1 & 63.0 & 21.5 & 18.7 & 44.2 & 75.2 & 63.9 & 42.4 & 16.2 & 17.8 & 18.0 & 79.5 & 76.5 & 36.1 & 78.0 & 76.6 & 78.9 \\
OCRNet-HRNet-W18s~\cite{OCRNet} & 73.6 & 59.0 & 62.3 & 62.9 & 52.8 & 20.1 & 18.5 & 33.7 & 61.0 & 54.6 & 29.5 & 10.9 & 11.8 & 8.8 & 69.8 & 69.3 & 30.4 & 70.0 & 69.5 & 72.4\\
ISANet (ResNet50)~\cite{ISANet} & 78.4 & 64.7 & 63.2 & 64.6 & 51.0 & 19.7 & 11.8 & 33.5 & 69.5 & 50.2 & 30.0 & 9.0 & 11.8 & 11.7 & 76.8 & 69.0 & 23.0 & 76.3 & 60.8 & 76.1\\
ISANet (ResNet101)~\cite{ISANet} & 79.6 & 67.4 & 67.6 & 67.6 & 56.2 & 20.0 & 19.8 & 37.1 & 75.1 & 55.1 & 33.9 & 10.6 & 13.8 & 14.2 & 78.2 & 72.8 & 28.9 & 77.7 & 65.3 & 76.9 \\
STDC (Pre-training)~\cite{STDC} & 75.0 & 64.5 & 65.1 & 64.9 & 54.7 & 21.8 & 11.3 & 32.2 & 68.9 & 51.1 & 29.0 & 26.4 & 10.9 & 6.4 & 73.2 & 67.9 & 31.1 & 73.1 & 61.0 & 73.2 \\  
STDC (No pre-training)~\cite{STDC} & 71.8 & 59.6 & 63.6 & 62.5 & 52.5 & 24.6 & 11.4 & 27.4 & 59.1 & 48.9 & 35.0 & 12.4 & 15.1 & 12.7 & 68.9 & 62.1 & 49.6 & 69.0 & 70.9 & 71.1\\ \midrule
SegFormer-b3~\cite{SegFormer} & 81.9 & 74.5 & 74.2 & 74.2 & 68.1 & 31.6 & 43.8 & 55.1 & 79.2 & 70.4 & 68.5 & 51.8 & 57.2 & 50.5 & 81.4 & 80.7 & 60.6 & 81.1 & 74.1 & 80.5 \\  
SegFormer-b0~\cite{SegFormer}  & 76.5 & 64.6 & 67.5 & 67.7 & 57.2 & 27.5 & 27.5 & 40.9 & 71.2 & 56.3 & 51.9 & 26.5 & 31.1 & 27.6 & 74.9 & 73.7 & 47.5 & 74.3 & 68.1 & 74.3 \\  \midrule
OneFormer-SwinTransfomer~\cite{OneFormer} & 83.0 & 79.8 & 78.0 & 77.4 & 73.9 & 35.2 & 65.9 & 56.8 & 81.6 & 78.8 & 77.8 & 67.5 & 72.7 & 73.6 & 82.5 & 82.0 & 71.8 & 82.5 & 77.4 & 81.7\\
OneFormer-ConvXNet~\cite{OneFormer} & 83.0 & 79.0 & 78.0 & 77.4 & 73.9 & 35.2 & 57.3 & 56.6 & 82.3 & 76.1 & 79.1 & 71.0 & 74.4 & 72.3 & 82.7 & 82.3 & 71.0 & 82.4 & 78.6 & 81.0\\ \midrule
SAM-SegFormer & 73.0 & 65.7 & 63.1 & 64.4 & 63.1 & 24.5 & 38.9 & 44.1 & 67.5 & 60.0 & 63.5 & 52.9 & 56.7 & 50.5 & 71.7 & 69.6 & 60.9 & 72.8 & 69.3 & 71.6\\
SAM-OneFormer & 80.0 & 75.5 & 73.7 & 72.3 & 70.8 & 29.5 & 58.3 & 51.7 & 76.9 & 72.4 & 74.0 & 63.9 & 68.7 & 67.6 & 78.9 & 77.2 & 68.9 & 79.5 & 72.7 & 77.8\\
MobileSAM-SegFormer & 68.9 & 61.7 & 59.0 & 60.0 & 58.9 & 20.9 & 32.1 & 37.4 & 60.7 & 53.7 & 57.5 & 47.7 & 51.4 & 44.9 & 67.5 & 61.9 & 56.3 & 68.5 & 65.5 & 67.6\\
MobileSAM-OneFormer & 75.3 & 70.5 & 68.9 & 67.5 & 66.5 & 25.7 & 44.2 & 44.6 & 69.7 & 64.0 & 67.0 & 56.5 & 61.1 & 58.5 & 73.9 & 68.5 & 63.3 & 74.6 & 68.5 & 73.1\\ \bottomrule
\end{tabular}
}
\end{table*}
\begin{table*}[t]\scriptsize
\centering
\setlength\tabcolsep{2pt}
\caption{Robustness of semantic segmentation models under the worst black-box corruptions (Severity=5).}
\label{table:black_box_corruption_severity5}
\resizebox{\textwidth}{!}{
\begin{tabular}{@{}l|c|ccccc|cccc|cccc|cccccc@{}}
\toprule
                 &      & \multicolumn{5}{c|}{\textbf{Blur}}         & \multicolumn{4}{c|}{\textbf{Weather}} & \multicolumn{4}{c|}{\textbf{Noise}} & \multicolumn{6}{c}{\textbf{Digital}}                        \\ \midrule
                 Architecture & \textbf{clean} & Gaussian & \textbf{defocus} & \textbf{motion} & glass & zoom & \textbf{snow}   & \textbf{frost}   & \textbf{fog}    & \textbf{spatter}   & speckle & Gaussian & shot & impulse & brightness & contrast & JPEG & saturate & pixelate & elastic \\ \midrule
SAM-SegFormer & 73.0 & 38.2 & 42.4 & 47.3 & 41.5 & 17.4 & 23.6 & 25.7 & 57.2 & 47.5 & 47.1 & 20.1 & 27.7 & 22.0 & 68.7 & 48.0 & 38.8 & 64.6 & 59.4 & 67.9\\
SAM-OneFormer & 80.0 & 58.2 & 59.9 & 55.6 & 53.6 & 19.4 & 36.0 & 34.2 & 70.4 & 63.3 & 62.0 & 26.0 & 30.4 & 32.2 & 76.6 & 62.4 & 48.1 & 71.7 & 13.4  & 73.1\\
MobileSAM-SegFormer & 68.9 & 35.6 & 39.8 & 42.7 & 38.0 & 14.1 & 16.0 & 19.7 & 54.1 & 24.2 & 40.8 & 15.7 & 22.0 & 15.6 & 68.7 & 38.8 & 35.5 & 59.7 & 56.4 & 63.9\\
MobileSAM-OneFormer & 75.3 & 52.9 & 55.4 & 50.8 & 50.3 & 16.0 & 21.0 & 25.0 & 66.3 & 31.4 & 67.0 & 20.4 & 25.7 & 25.4 & 76.6 & 49.7 & 43.8 & 65.9 & 13.2 & 69.1\\ \bottomrule
\end{tabular}
}
\end{table*}
\par Autonomous driving systems will face various perturbations in the complex environment of the real world, including equipment noise, drastic changes in light, adverse weather, and so on. The corrupted image modal data will affect the recognition performance of the system, and it is essential to have a systematic study on the robustness of the perception system under such black-box corruption scenarios. The corrupted Cityscapes~\cite{cityscapes} dataset with corrupted images covers a total of 19 types of image corruption in five major categories, including noise, blur, adverse weather, and digital disturbance. This subsection comprehensively evaluates the robustness of the black box corruption based on the SAM model.
\par \textbf{It is important to emphasize that the SAM model is not trained additionally on the Cityscapes dataset~\cite{cityscapes}, and all inferences are conducted in a zero-shot manner.} As a comparison, all other models have undergone training on the Cityscapes dataset, whereas the SAM models have not exploited the training set of Cityscapes. The severity level is set at the Level 1, Level 2, and Level 3. The metric of $m I o U$ is calculated as the average value on these three levels. Table \ref{table:black_box_corruption} illustrates the results, and the bolded black box attack items are related to SOTIF. The performance of the SAM model exceeds that of most CNN-based models and even a small subset of Transformer-based models. It demonstrates the effectiveness and potential of the SAM model, even without any additional training on the Cityscapes dataset. Compared with the SegFormer model and the OneFormer model, which are fully trained on the Cityscapes dataset, the gap is not large. Overall, the recognition performance of the vanilla SAM model exceeds the MobileSAM model at the cost of inference speed and storage space.
\par Table \ref{table:black_box_corruption_severity5} illustrates the robustness of the SAM models under extremely adverse conditions when the black-box corruption severity is set to 5. For example, the robustness of SAM-OneFormer is remarkable under such black-box attacks, including Gaussian blur, defocus blur, motion blur, glass blur, snow, frost, fog, spatter, speckle noise, shot noise, impulse noise, brightness, contrast, JPEG compression, and elastic transform. However, the robustness of SAM-OneFormer under black-box attacks with Gaussian noise and pixelate is not considerable.
\subsubsection{Robustness under the FGSM Attacks}

\begin{table}[!t]
\centering
\caption{Robustness (mIoU) of different semantic segmentation models under the FGSM attacks ($\varepsilon=8.0/255.0,16.0/255.0$).}
\vspace{0.1cm} 
\begin{tabularx}{\linewidth}{lcc}
\toprule
 &  $\varepsilon=8.0/255.0$ & $\varepsilon=16.0/255.0$ \\
\midrule
DP-ResNet50~\cite{Deeplabv3+} & 36.9 & 16.6 \\
DP-ResNet101~\cite{Deeplabv3+} & 44.1 & 19.3 \\
DP-MobileNetV2~\cite{Deeplabv3+} & 30.6 & 10.8 \\
DP-Xception65~\cite{Deeplabv3+} & 17.1& 4.9 \\
FCN32s-ResNet50~\cite{FCN} & 15.4 & 3.3 \\
FCN32s-ResNet101~\cite{FCN} & 26.6 & 10.0 \\
FCN32s-VGG16~\cite{FCN} & 14.3 & 7.2 \\
FCN16s-VGG16~\cite{FCN} & 10.3 & 6.8 \\
FCN8s-VGG16~\cite{FCN} & 12.2 & 6.5 \\
PSPNet-ResNet50~\cite{Pspnet} & 12.3 & 6.0 \\
PSPNet-ResNet101~\cite{Pspnet} & 18.3 & 3.9 \\
SegNet-VGG16~\cite{Segnet} & 22.9 & 10.8 \\
STDC (Pre-training)~\cite{STDC} & 9.8 & 2.0 \\
STDC (No pre-training)~\cite{STDC} & 11.7 & 4.8 \\
SegFormer\_mit-b5~\cite{SegFormer} & 56.6 & 49.2 \\
SegFormer\_mit-b3~\cite{SegFormer} & 49.8 & 37.2 \\
SegFormer\_mit-b0~\cite{SegFormer} & 35.3 & 22.0 \\
OCRNet-ResNet100~\cite{OCRNet} & 11.6 & 1.7 \\
OCRNet-HRNet-W48~\cite{OCRNet} & 30.3 & 3.5 \\
OCRNet-HRNet-W18~\cite{OCRNet} & 11.6 & 2.4 \\
ISANet (ResNet50)~\cite{ISANet} & 13.9 & 3.2 \\
ISANet (ResNet101)~\cite{ISANet} & 29.7 & 8.1 \\
SAM-SegFormer & 51.6 & 44.8 \\
SAM-OneFormer & 59.1 & 57.6 \\
MobileSAM-SegFormer & 49.4 & 43.0 \\
MobileSAM-OneFormer & 56.3 & 52.9 \\
\bottomrule
\end{tabularx}
\label{table:fgsm_miou_values}
\end{table}
 \begin{figure}[!t]
        \centering
\includegraphics[width=0.8\hsize]{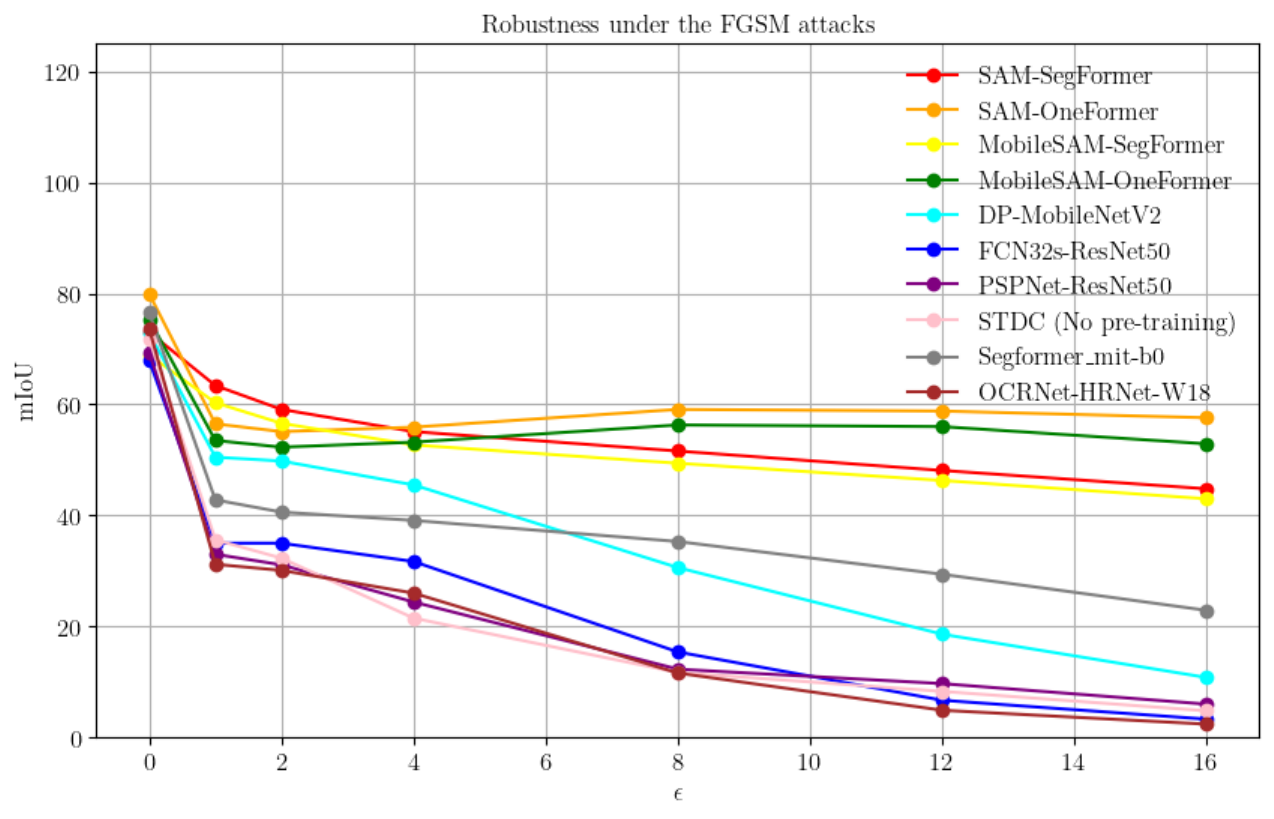}
\caption{Robustness study under the FGSM attacks}
\label{fig:sam_fgsm}
\end{figure}
\par Table \ref{table:fgsm_miou_values} illustrates the robustness of the semantic segmentation models under the single-step FGSM attacks. Although the record of the robust mIoU of OneFormer is not recorded in this paper, the aim of this study is not the announcement of SAM architectures as state-of-the-art robust segmentation models. This study focuses on the exploration of the zero-shot adversarial robustness of SAM in autonomous driving. The SAM models maintain zero-shot adversarial robustness under the FGSM attacks, and their performance even surpasses that of many supervised learning models.
\par Fig. \ref{fig:sam_fgsm} illustrate the robustness with the different perturbation budgets $\varepsilon$ in the range of $[1.0/255.0,\ 2.0/255.0,\ 4.0/255.0,\\ \ 8.0/255.0,\ 12.0/255.0,\ 16.0/255.0]$. All of the SSAM variants, including SAM-SegFormer, SAM-OneFormer, MobileSAM-SegFormer, and MobileSAM-OneFormer, realize adversarial robustness under the FGSM attacks. Furthermore, the robust mIoUs of SAM-OneFormer and MobileSAM-OneFormer even achieve a slight rise under the larger perturbation budget. The potential reason is the loss of efficacy of the single-step gradient-based attack on the combination paradigm of SAM and OneFormer.

\subsubsection{Robustness under the PGD Attacks}
\begin{table}[!t]
\centering
\caption{Robustness (mIoU) of different semantic segmentation models under the PGD-10 attacks ($\varepsilon=8.0/255.0,16.0/255.0$).}
\vspace{0.1cm} 
\begin{tabularx}{\linewidth}{lcc}
\toprule
 &  $\varepsilon=8.0/255.0$ & $\varepsilon=16.0/255.0$ \\
\midrule
PSPNet~\cite{Pspnet} & 28.8 & 26.0 \\
DeepLabV3~\cite{Deeplabv3} & 29.5 & 26.5 \\
SAM-SegFormer & 21.6 & 21.3 \\
SAM-OneFormer & 53.5 & 52.1 \\
MobileSAM-SegFormer & 19.8 & 20.6 \\
MobileSAM-OneFormer & 49.6 & 52.1 \\
\bottomrule
\end{tabularx}
\label{table:pgd_miou_values}
\end{table}
 \begin{figure}[!t]
        \centering
\includegraphics[width=0.8\hsize]{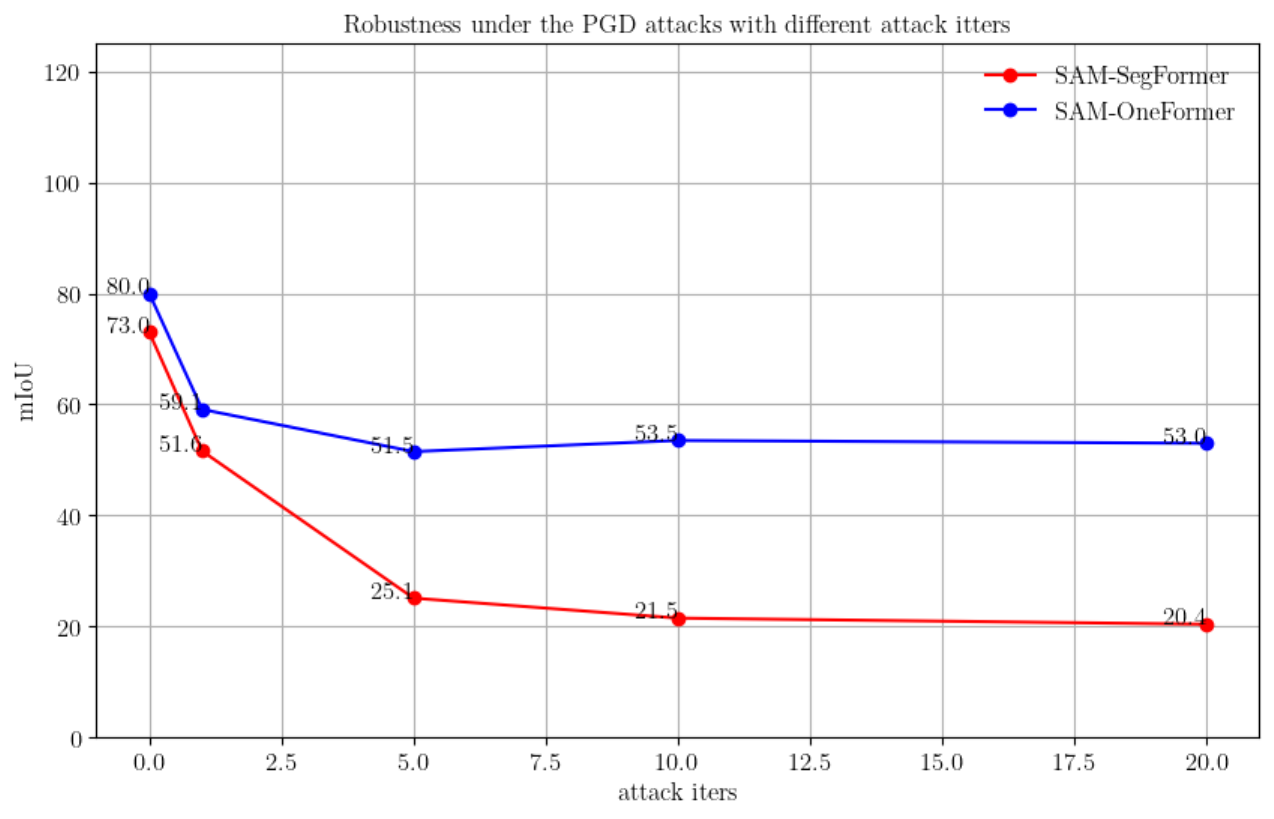}
\caption{Robustness study under the PGD attacks}
\label{fig:sam_pgd}
\end{figure}
\par This subsubsection focuses on the robustness study of SAM under the PGD attacks. The PGD attack is an authoritative evaluation method in robustness assessment for gradient-based attacks. In the PGD attack, the parameter $\alpha$ typically represents the step size for updating the input sample at each iteration. Table \ref{table:pgd_miou_values} illustrates the robustness of the semantic segmentation models under the PGD attacks with 10 attack iterations. We test the scenarios under the perturbation budgets $\varepsilon$ in the range of $[ 8.0/255.0,\ 16.0/255.0]$. The record of PSPNet~\cite{Pspnet} and DeepLabV3~\cite{Deeplabv3} are quoted from the previous study~\cite{SegPGD}. The experimental results demonstrate the zero-shot adversarial robustness of visual foundation models in the semantic segmentation task. Moreover, this subsection explores the influence of attack iteration besides the influence of perturbation budget. Fig. \ref{fig:sam_pgd} shows the influence of attack iterations. We test SAM-SegFormer and SAM-OneFormer. It shows that both models experience a drop in the mIoU metric as the number of attack iterations increases, indicating that their performance degrades under PGD attacks. However, SAM-OneFormer appears to be more robust, maintaining higher mIoU values compared to SAM-SegFormer across all attack iterations.
\subsection{Discussion}
\begin{table}[!t]
\centering
\caption{Comparison between SAM and MobileSAM}
\vspace{0.1cm} 
\begin{tabular}{lcc}
\toprule
 & Params (M) & Inference Speed (ms) \\
\midrule
SAM~\cite{SAM} & 632 & 452 \\
MobileSAM~\cite{MobileSAM} & 5.78 & 8 \\
\bottomrule
\end{tabular}
\label{table:compare}
\end{table}
\par \textbf{Zero-shot adversarial robustness of Semantic-Segment-Anything models in autonomous driving:} The experimental results demonstrate that SAM and MobileSAM can realize adversarial robustness under the white-box attacks and black-box attacks. The robustness guarantee is more obvious on the OneFormer backbone. The robustness under the black-box corruptions connects the adverse conditions like bad weather and sensor noises in autonomous driving, which is significant to the SOTIF of autonomous driving. Moreover, the robustness under the white-box attacks is important to the security of the perception systems in the Internet of Vehicles. The phenomenon of zero-shot adversarial robustness would inspire the work for the further improvement of internal safety and external security in autonomous driving. Furthermore, the change of model design paradigm brings about a change in the robustness study paradigm. The malicious hackers may propose a new attack method adapted to the SAM variants.
\par \textbf{Trade-off between robustness and cost:} Overall, the robustness of SAM exceeds the robustness of MobileSAM on the Cityscapes dataset. However, Table \ref{table:compare} demonstrates the cost of storage and inference of the SAM architectures in the platform with the GeForce RTX 3090 GPU. Although MobileSAM does not have the highest robustness, it is more suitable for computation on edge devices.
\par \textbf{The robustness difference between the SegFormer and OneFormer backbones:} Both SegFormer and OneFormer are based on self-attention mechanisms. SegFormer uses a hierarchical Transformer encoder and a lightweight MLP decoder capable of outputting multiscale features without positional encoding. OneFormer unifies semantic, instance, or panoramic segmentation tasks, and the model is designed as task-dynamic. OneFormer can dynamically adjust its outputs according to the task and category tokens during inference, thus enabling category-specific and mask-specific predictions. This inference mechanism of OneFormer may mitigate the adversaries' gradient perturbation.
\par \textbf{The limitations of current framework:} Some new strong attack methods~\cite{MDAttack} may launch more severe threats on the ViT backbones. The current proposed method does not integrate the sufficient test-time defense module. Moreover, the test scale can be enlarged.

\section{Conclusion}
\par This paper explores the zero-shot adversarial robustness of SAM architectures in the semantic segmentation task for autonomous driving. The findings are surprising, showing that this type of model exhibits robustness under black-box attacks related to adverse weather and sensor interference. These results provide valuable insights into the Safety of the Intended Functionality (SOTIF) of autonomous driving systems. Additionally, the model demonstrates considerable robustness against white-box adversarial attacks, offering a security guarantee against malicious data in the Internet of Vehicles.
\par In the future, we plan to expand the test scale to include other attack methods such as SegPGD~\cite{SegPGD}. Moreover, integrating test-time defense methods to further enhance robustness is a meaningful direction for research. Thirdly, the interpretation of the zero-shot adversarial robustness of visual foundation models remains an open area that requires further exploration. Last but not least, the study of deployment in real-world applications deserves further study to build a next-generation trustworthy AGI system.
\section*{Acknowledgments}
This work was supported by the Shanghai International Science and Technology Cooperation Project No.22510712000 and the Special Funds of the Tongji University for ``Sino-German Cooperation 2.0 Strategy" No. ZD2023001. The authors would like to thank T{\"{U}}V S{\"{U}}D for the kind and generous support. We are also grateful for the efforts from our colleagues in Sino German Center of Intelligent Systems. We are sincerely grateful for the support from Mr. Ronghui Liu for the software implementation.

\bibliographystyle{IEEEtran}
\bibliography{references.bib}
\vspace{12pt}
\color{red}

\end{document}